\documentclass[lettersize,journal]{IEEEtran}
\usepackage{cite}

\usepackage{amsmath,amsfonts}
\usepackage{algorithmic}
\usepackage{array}
\usepackage[caption=false,font=normalsize,labelfont=sf,textfont=sf]{subfig}
\usepackage{textcomp}
\usepackage{stfloats}
\usepackage{url}
\usepackage{verbatim}
\usepackage{graphicx}
\def\BibTeX{{\rm B\kern-.05em{\sc i\kern-.025em b}\kern-.08em
    T\kern-.1667em\lower.7ex\hbox{E}\kern-.125emX}}
\usepackage{balance}
\begin{document}
\title{Integrated feature analysis for deep learning interpretation and class activation maps}
\author{Yanli Li, Tahereh Hassanzadeh, Denis P. Shamonin, Monique Reijnierse, Annette H.M. van der Helm-van Mil and Berend C. Stoel
\thanks{This work is supported by the Netherlands Organization for Scientific Research (NWO, TTW 13329), the ERC (European Research Counsel) starting grant under the European Union’s Horizon 2020 research and innovation programme No.714312 and the China Scholarship Council No.202108510012.}
\thanks{Yanli Li, Tahereh Hassanzadeh, Denis P. Shamonin and Berend C. Stoel are with the Division of Image Processing, Department of Radiology, Leiden University Medical Center, (LUMC), 2300 RC, Leiden, The Netherlands.}
\thanks{Monique Reijnierse is with the Department of Radiology, Leiden University Medical Center, (LUMC), 2300 RC, Leiden, The Netherlands.}
\thanks{Annette H.M. van der Helm-van Mil is with the Department of Rheumatology, Leiden University Medical Center, (LUMC), 2300 RC, Leiden, The Netherlands.}
}
\maketitle

\begin{abstract}
Understanding the decisions of deep learning (DL) models is essential for the acceptance of DL to risk-sensitive applications, such as security systems, industrial anomaly detection and medical imaging. Although methods, like class activation maps (CAMs), give a glimpse into the black box, they do miss some crucial information, thereby limiting its interpretability and merely providing the considered locations of objects. To provide more insight into the models and the influence of datasets, we propose an integrated feature analysis method, which consists of feature distribution analysis and feature decomposition, to look closer into the intermediate features extracted by DL models. This integrated feature analysis could provide information on overfitting, confounders, outliers in datasets, model redundancies and principal features extracted by the models, and provide distribution information to form a common intensity scale, which are missing in current CAM algorithms. The integrated feature analysis was applied to eight different datasets for general validation: photographs of handwritten digits, two datasets of natural images and five medical datasets, including skin photography, ultrasound, CT, X-rays and MRIs. The method was evaluated by calculating the consistency between the CAMs average class activation levels and the logits of the model. Based on the eight datasets, the rescaled CAMs achieved on average increases in consistency of around 23.1\%, 51.6\%, 15.5\%, 22.7\%, 32.9\%, 10.7\%, 64.2\% and 17.1\%, respectively. The correlation coefficients were all very close to 100\%, proving the effectiveness of the distribution analysis in the integrated feature analysis. Moreover, based on the feature decomposition, 5\%-25\% of features could generate equally informative saliency maps and obtain the same model performances as using all features. This proves the reliability of the feature decomposition. As the proposed methods rely on very few assumptions, this is a step towards better model interpretation and a useful extension to existing CAM algorithms. Codes: \url{https://github.com/YanliLi27/IFA}
\end{abstract}


\section{Introduction}
\begin{figure*}[!t]
\centering
\includegraphics[width=\textwidth]{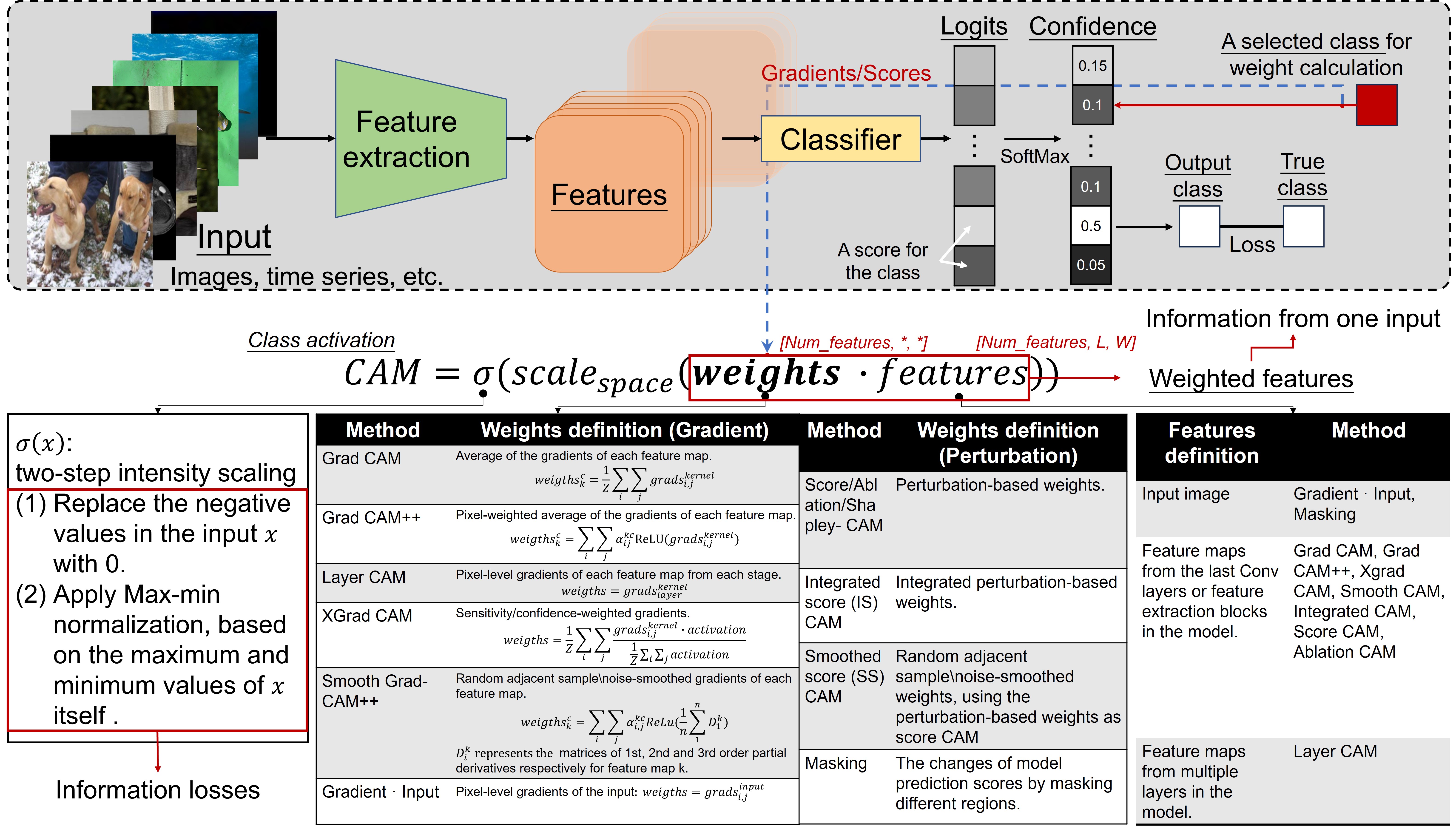}
\caption{Summary of the workflow, terms and formula involved in generating CAMs. The flowchart at the top shows the general DL workflow, including (1) inputting the \textit{input} to the feature extraction module to get the \textit{features}, (2) feeding the features to the classifier to get the \textit{logits}, (3) with the activation function like softmax to get the \textit{confidence}, and (4) based on the \textit{confidence} to get the \textit{output class} and compare with the ground truth \textit{true class}. The formula indicates the process of CAM algorithms to obtain features, the source of weights for the CAM calculation, the size for spatial scaling and the two-step intensity scaling that discards the negative values and normalizes the input based on the input itself. The table at the bottom gives a summary of the weight calculations of the existing CAM methods and the chosen features. Red boxes highlight where information is lost or limited in current algorithms. Underlined terms indicate the following context, including \textit{input}, \textit{features}, \textit{logits}, \textit{confidence}, \textit{output class}, \textit{true class}, the \textit{selected class} for CAM generation, \textit{class activation} for describing the class activation level of CAM and \textit{weighted features}. In existing class activation mapping algorithms, the selected class is set by default to the output class.}
\label{Fig. 1}
\end{figure*}

\IEEEPARstart{P}{roviding} insight into how deep learning (DL) models make decisions is a prerequisite before DL models can be applied in risk-sensitive areas. The potential risk of making decisions based on confounders is a non-trivial issue in risk-averse domains like healthcare \cite{future_and_risk}, security systems \cite{AIrisk} and other practices \cite{AIinIndustry}.
\noindent An intuitive idea to validate the reliability of deep learning models is to confirm that models are making decisions based on reasonable evidence and convincing features. Since there is no formal definition of interpreting DL models, current interpretation algorithms rely on some commonly agreed principles \cite{Interpret}. For instance, a dog should be classified based on the dog’s face instead of the background or other objects in the image, or a bird should be distinguished from other animals because of its wings. Three principles are mainly followed in current studies: (1) analyzing the responses of or changes in the model’s output to investigate the model, by assessing the individual contributions of samples in datasets \cite{inchange1}, estimating the learning difficulty \cite{inchange2} or detecting misclassified samples \cite{inchange3}; (2) analyzing low-level features extracted by the models to investigate the model’s feature extraction process \cite{rationale1,CAM,rationale3}; and (3) analyzing the contribution of certain parts in the input image to the model’s output, evaluating whether the contribution of these parts is consistent with human knowledge. Compared to some early efforts based on principles (1) and (2), methods based on principle (3) are more intuitive and easier for visual checks and understanding.
\noindent As a typical representative of methods based on principle (3), class activation mapping and its variants, which generate class activation maps (CAMs), demonstrate high computational efficiency compared to other widely-used methods, such as LIME and its variants \cite{Anchors, SHAP, RISE, MAPLE}, global interpretation \cite{limesp, normlime, gale} and other interpretation methods \cite{LRP, Start1, Start2, Start3, Start4, Start5}, becoming one of the most popular methods for investigating model reliability and interpretability.  
\noindent CAMs (also known as saliency maps, heatmaps or attention maps) provide information on the locations of objects considered by the given models, based on the assumption that weighted features could precisely describe the model outputs. The weighted features in CAM algorithms usually refer to the summed features from a certain layer of the models, weighted by the designed weights that are assumed to represent the “importance” of these features for the final output class. Fig. \ref{Fig. 1} presents a summary of the workflow and terms used in generating CAMs and the formulas for the most common variations of CAMs, including the original CAM \cite{CAM}, Grad CAM \cite{GradCAM}, Grad CAM++ \cite{GradCAMpp}, Layer CAM \cite{LayerCAM}, xGrad CAM \cite{XGradCAM}, Score CAM \cite{ScoreCAM}, Smooth Grad CAM++ \cite{SmoothGradCAMpp}, SS-CAM \cite{SSCAM} and IS-CAM \cite{ISCAM}, and a flowchart of the shared process. These algorithms gave promising performances on complicated tasks by locating the target objects in the images with complex backgrounds and foregrounds \cite{WSOLDsurvey2} and developed the field of weakly supervised object detection \cite{WSOLDsurvey}.

\begin{figure}[!t]
\centering
\includegraphics[width=\linewidth]{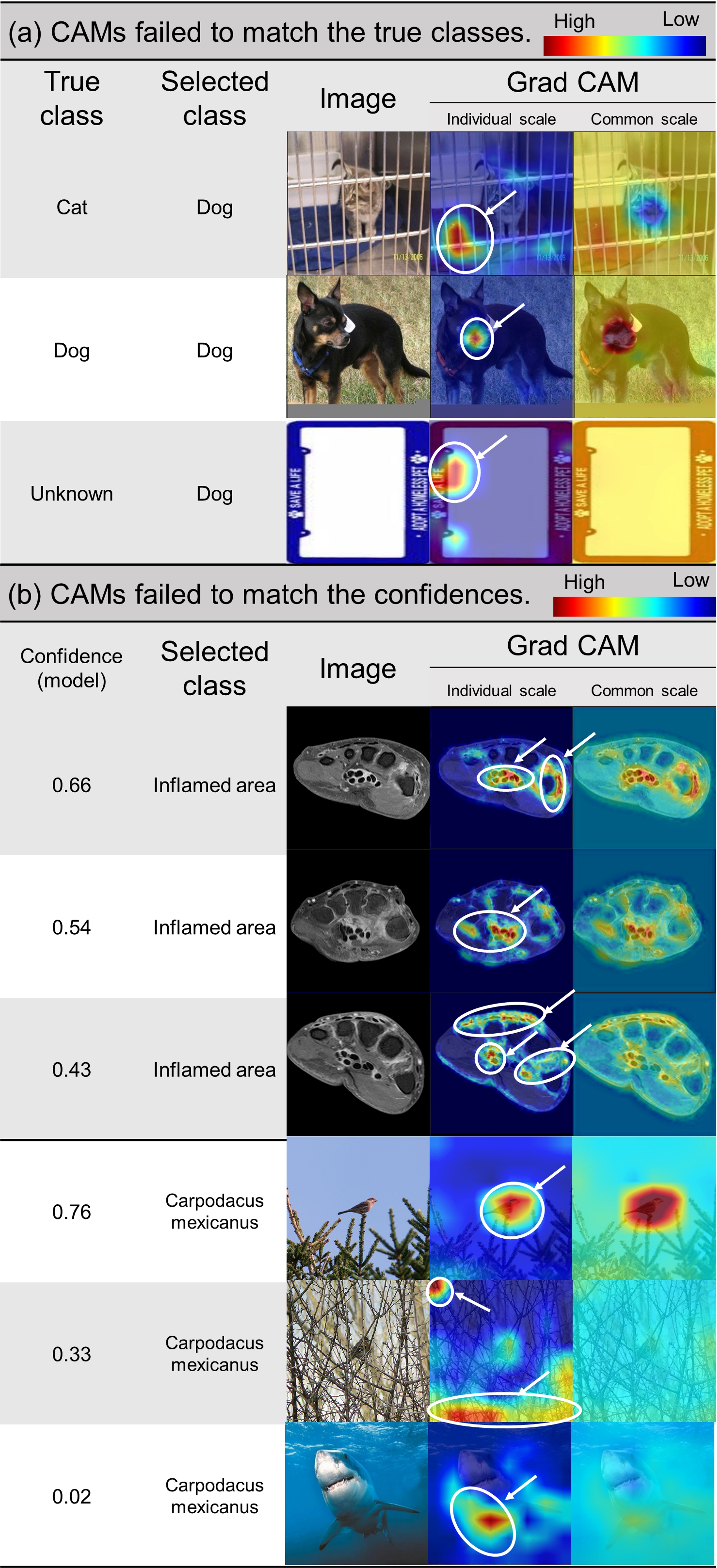}
\caption{Consequences of omitting scale information. (a) CAMs fails to match the true class. In the CAMs for the selected class ‘dog’, an image of a ‘cat’ receives higher class activations than a correctly classified image of ‘dog’. In addition, unknown inputs still receive considerable class activations in the CAMs by a model that can distinguish cats and dogs with a 99\% accuracy; (b) CAMs failed to match the confidence in the models trained for an MRI-based classification task and ImageNet.}
\label{Fig. 2}
\end{figure}

\noindent Although these algorithms provide important insight, some important information is lost in the calculation process as shown in the red boxes in Fig. \ref{Fig. 1}: (1) due to intensity scaling at a single-image level – an individual intensity scale, real (unscaled) class activation levels are missing of each input image, which would be useful for the comparison with other input images, are missing due to intensity scaling at a single-image level – an individual intensity scale; (2) pixels that provide negative contributions (lowering the confidence based on some reasonable hints) to the logits of the selected classes are dismissed by selecting only positive class activations to create CAMs in current methods; and (3) features that are very frequently activated for some classes over the whole dataset are treated the same as features that are hardly activated at all, since CAMs simply sum the weighted features together.
\noindent As a result, the only information left in current CAMs is the considered location of the objects in one image, which can be misleading. Consequently, these CAMs may be inconsistent with respect to the corresponding logits and confidence of the model. Some images that did not contain target objects of the selected class still received higher class activations in the CAMs than those with the target objects (See Fig. \ref{Fig. 2}(a)). Moreover, without ground truths or prior knowledge, it is impossible to determine from the CAMs when the model has high confidence in the output class or when the model is struggling to classify correctly (See Fig. \ref{Fig. 2}(b)). Confounders may have falsely “contributed” to the model outputs, yet they are invisible because negative class activations are discarded during the generation of CAMs. Furthermore, the lack of presenting more detailed information in CAMs could limit its applications to the fields that require scalar outputs rather than binary outputs, like the severity of diseases or lesions in some specific regions in medical imaging.
\noindent Generally, current algorithms do not provide information on the working of a model, apart from the location of the target objects of the selected classes. Providing additional information about the models and the datasets requires methods to be developed, based on principles (2) and (3).

\noindent Based on the same assumption of utilizing weighted features, we aimed to retrieve more information by providing more insight into the trained models and datasets, and solving the above problems of current algorithms. Specifically, we propose an integrated feature analysis that consists of a distribution analysis and a feature decomposition, focusing on the statistical information (e.g., peaks, valleys and spread of the weighted features) and the independence of each weighted feature (e.g., frequency of becoming activated over the whole dataset), respectively. Visualized and validated by modifying and improving the CAMs, the integrated feature analysis enabled us to fill the gaps between models and CAMs and improve the understanding of models and datasets similar to the method based on principles (2) and (3).
\noindent The layout of this paper is as follows. First, we introduce two parts of the proposed integrated feature analysis: (1) distribution analysis and (2) feature decomposition, followed by the CAM generation process based on these two parts. Subsequently, we introduce the methodology to evaluate the proposed method. In the section on evaluation, some quantitative evaluation methods are introduced, including current evaluation metrics based on object localization and a new method, based on consistency that excludes the impact of model performances and prior knowledge involvements. We validated the integrated feature analysis by proving that distribution analysis could help to improve the consistency between the class activation levels of CAMs and the corresponding model confidence and logits, and feature decomposition contributes to finding the principal features that contain the main information for the model outputs. Following the methodology and evaluation, we briefly introduce the eight datasets from different subjects and modalities, used for validation. After the basic information, we present the overall performance of the proposed methods on these eight datasets. In the last two chapters, we discuss and summarize some applications that could improve the understanding of models and datasets, together with the advantages and limitations of the proposed method.

\section{Method}
\begin{figure*}[!t]
\centering
\includegraphics[width=\textwidth]{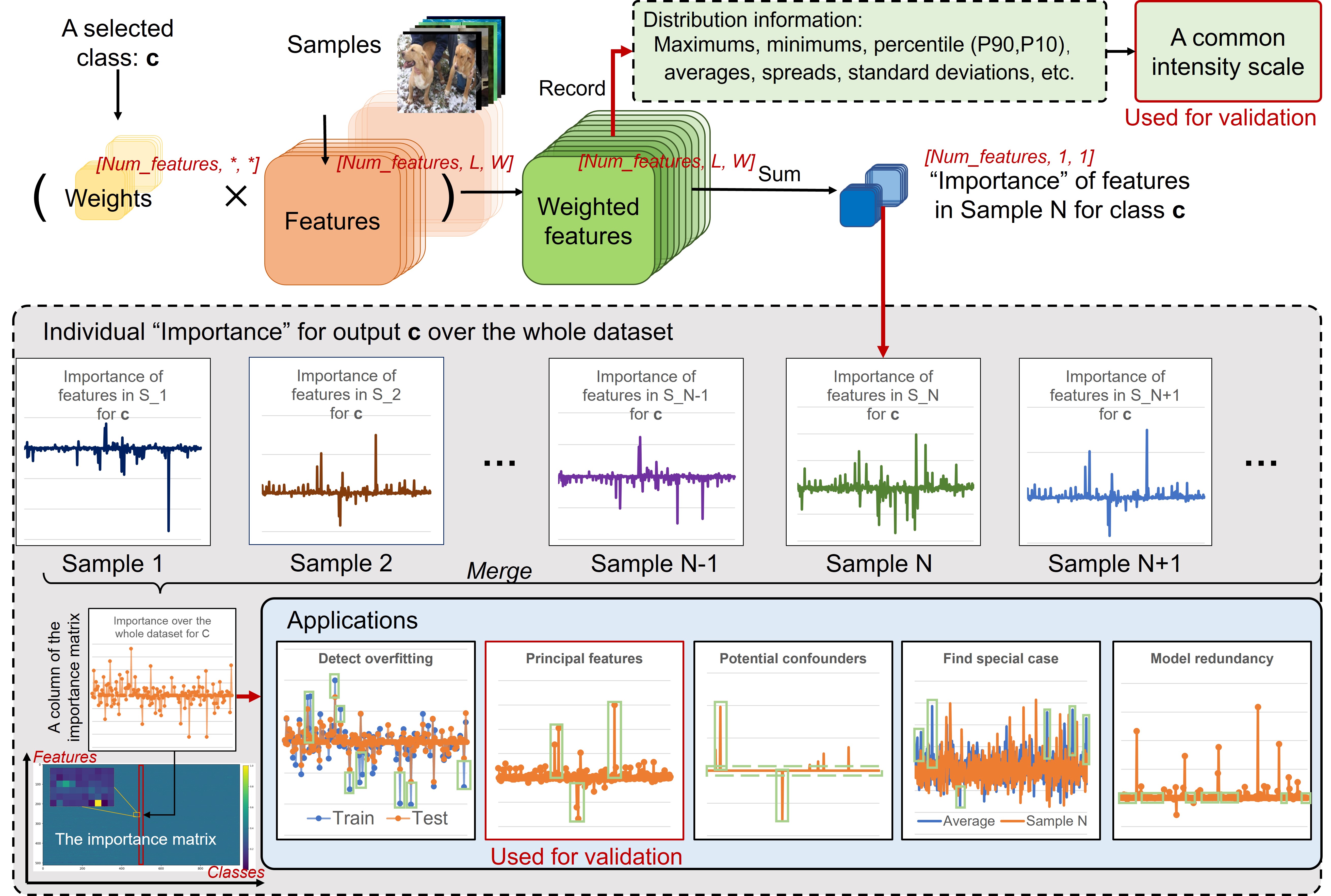}
\caption{Process of integrated feature analysis that collects information about the distribution of features across the dataset and computes the importance matrix for feature decomposition. The x-axis in the importance matrix represents different classes, while the y-axis represents different features at the selected level of a model. The box presents some applications of the importance matrix, including detecting overfitting, determining principal features, finding potential confounders, locating special cases, and evaluating model redundancies by analyzing the class activations in the importance matrix. Further details are provided in the Discussion section.}
\label{Fig. 3}
\end{figure*}

\noindent The integrated feature analysis consists of two parts: (1) distribution analysis and (2) feature decomposition. The distribution analysis aims at determining the spread, standard deviation and percentiles of the class activation levels of weighted features, in order to compare these levels from one sample to another or to the average level over the whole dataset. The feature decomposition focuses on the value of each feature through a so-called “importance matrix”. Fig. \ref{Fig. 3} presents the process of the integrated feature analysis and the calculation of the importance matrix, including the path for distribution analysis before summing up the weighted features and the path for calculating a column of the importance matrix (with a shape of [number of features ($F$), number of output classes ($C$)]) for the selected class $c$.

\subsection{Distribution analysis}
\begin{figure}[!t]
\centering
\includegraphics[width=\linewidth]{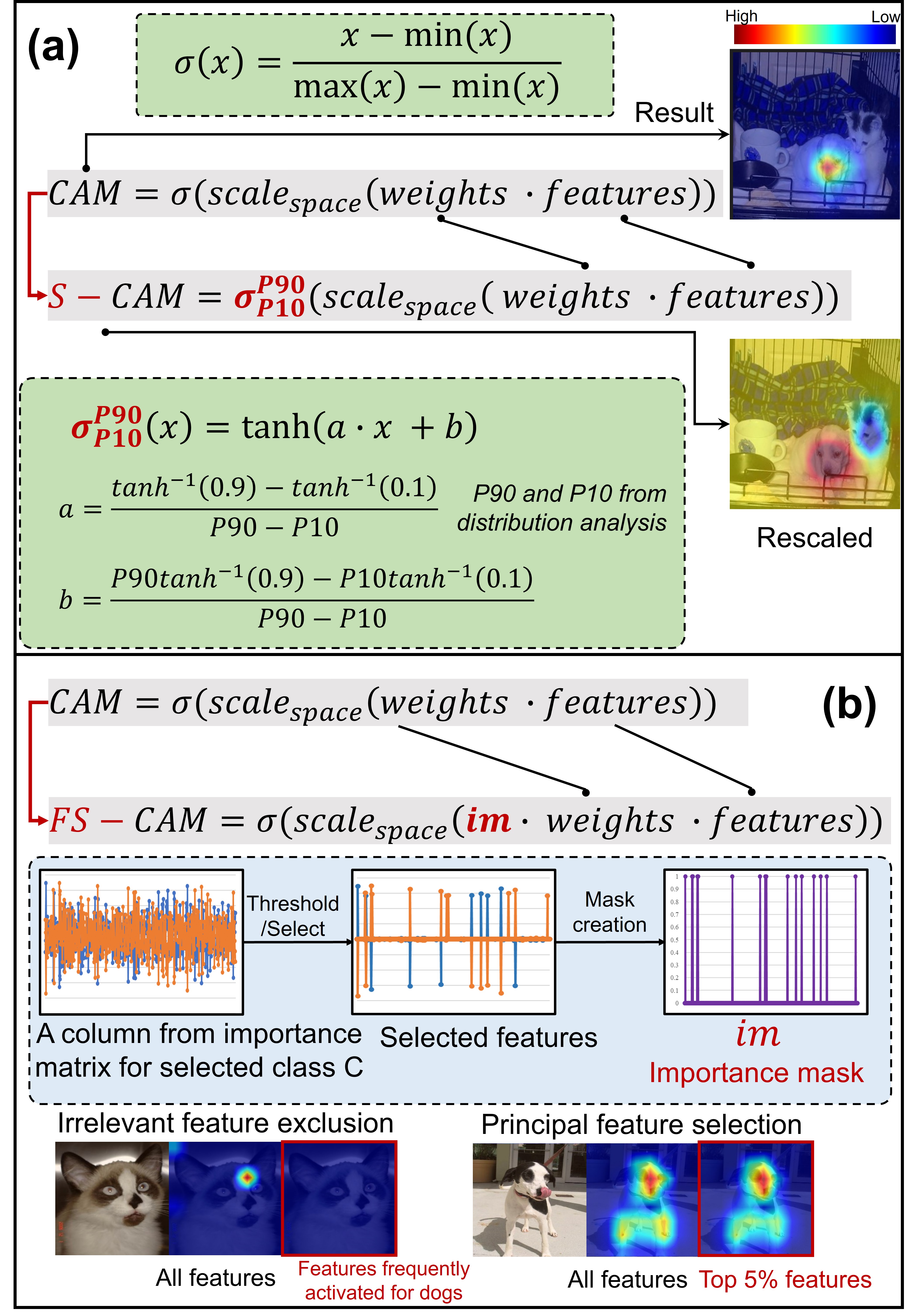}
\caption{CAM generation with integrated feature analysis: (a) distribution information to improve the intensity scaling. After the intensity scaling, the cat in the image gains a lower contribution compared to the background, which is intuitive for a cat-dog classification model; (b) importance matrix to obtain CAMs for specific purposes. The bottom images present CAMs with the purpose of excluding irrelevant features and selecting principal features.}
\label{Fig. 4}
\end{figure}

\noindent The analysis of the feature distribution forms the basis for feature decomposition and standard CAM generation. The collected information enables a common intensity scale for the CAM generation (shown in Fig. \ref{Fig. 4}(a)), which could improve the consistency between the CAMs and the models.

\noindent The CAM calculation with a common intensity scale can be described as: instead of scaling the products of weights and features based on the maximum and minimum of only the current input, the whole dataset is used to scale the weights and features (training set if time-available, validation set if time-limited) obtained during the distribution analysis – a common intensity scaling compared to the individual intensity scaling in current methods. The revised saliency map generation is given by:
\begin{equation}\label{eq1}
\begin{split}
S-CAM=\sigma_{P10}^{P90} (scale_{space}&(weights \\
&\cdot features))\\  
\end{split}
\end{equation}
\noindent where $weights$ represents the weights designed by the CAM algorithms, $features$ refers to the chosen features, $scale_{space}$ is the spatial resizing function, and $P90/P10$ are the upper and lower limit for normalization. In this work, the $\sigma_{P10}^{P90}()$ is designed to be:
\begin{equation}\label{eq2}
\sigma_{P10}^{P90}(x)=tanh(\alpha \cdot x + \beta) 
\end{equation}
\noindent where $\alpha$ and $\beta$ are constants given by $\frac{tanh^{-1}0.9-tanh^{-1}0.1}{P90-P10}$ and $\frac{P90tanh^{-1}0.9-P10tanh^{-1}0.1}{P10-P90}$.
\noindent This scaling function with a hyperbolic tangent function is to prevent information loss outside the interval P10-P90, as would be the case in classical maximum-minimum normalization with clipping. At the same time, the central curves and the domain of definition of this function also enables the common intensity scale to amplify the differences in smaller class activation levels and adapt to external data input for more general uses.
\noindent Since the proposed intensity scaling and preservation of negative values can be considered a functional extension to existing methods, we preferred not to call it as a new CAM method, but rather an intensity-scaled version of these methods, labeled with “S-”. For example, for Grad CAM, after the proposed common intensity scaling, we call this “S-Grad CAM”.

\subsection{Feature decomposition}
\noindent The feature decomposition aims at determining the independence and importance of each feature, without considering the spatial information and the correlation with other features. We define the importance of a feature by the average contribution of this feature (measured through the sum of a weighted feature) to the model logits of the selected class over the whole dataset, as each feature represents a fixed feature extraction path after training the model (See Fig. \ref{Fig. 3}, the “Importance” of features). To determine the principal features, we introduce an importance matrix, measuring the importance of each feature based on the above definition of importance. 
\noindent Importance matrix is a measurement with a shape of [number of features ($F$), number of classes ($C$)] for the analysis of each feature, in which each row is the average class activation of a certain feature for all classes on the whole dataset, and each column indicates the average class activation of all features for a specific selected class. The $IM^{f, c}$ refers to the average class activation level of the $f$th feature for the $c$th class on the whole dataset. Instead of summing up the $F$ weighted features to generate the CAMs, we obtain the $F$ values for each input, model and the selected class $c$, by calculating the sums of each separate feature and regard these $f$th value in the $F$ values as the “importance” of the $f$th feature. As the model does not change and therefore the size of $F$ does not change, the corresponding $F$ values from different inputs can be accumulated. Finally, a vector with a shape of [$F$, 1($c$)] is obtained, containing $F$ values for each selected class that represents the average importance of each feature for each selected class on the whole dataset. By merging all vectors in the order of the features, the importance matrix of [$F$, $C$] can be obtained.
\begin{equation}
\label{eq3}
IM^{f, c} = (\sum_{n=0}^{N}(weight_{n}^{f, c} \cdot feature_{n}^{f, c}) )/N 
\end{equation}
\noindent Eq. \ref{eq3} defines a column of the importance matrix of the $f$th feature for a selected class $c$, where $N$ represents the total number of samples in the datasets and $n$ represents the $n$th sample, $f$ refers to the $f$th feature out of all features from the target layer in the deep learning model, $weight_{n}^{f, c}$ and $feature_{n}^{f, c}$ refers to the weights and features designed by the CAM algorithms for a selected class $c$ and $f$th feature out of all features.
\noindent Generating importance matrices based on Eq. \ref{eq3} is time-consuming if the models are used for datasets with a large number of classes, because the weights of CAMs are calculated based on a selected class, which requires to change the selected classes for calculation.  For example, for ImageNet 2012 that contains 1000 classes to be selected, this process would require more than 1000 for-loops to complete the importance matrix for all classes and is unacceptable.
\noindent To improve the efficiency for models with many classes, we therefore proposed to generate a unified importance matrix for all classes based on the true classes instead of generating importance matrices for each class. Therefore, the CAMs generated based on the gradients from output classes naturally have the highest class activation levels in the classification based on the Argmax function. For a model designed for a 1000-class classification task with 512 features at the selected layers of the model, this approach produces an importance matrix with a shape of [512, 1000] through only one loop. 
\begin{equation}\label{eq4}
IM^{f,c} = \frac{
{\textstyle \sum_{n=0}^{N}}  \left\{\begin{matrix} 
(weight_{n}^{f,c} \cdot feature_{n}^{f,c}) & \textit{gt=c}
 \\ 
0 &\textit{gt!=c} 
\end{matrix}\right.
}{N-num_{gt!=c}} 
\end{equation}
\noindent Eq. \ref{eq4} presents the modified calculation process, where $gt$ refers to the true classes of the inputs, $num_{gt!=c}$ represents the number of samples that belong to other classes, and others remain the same as the Eq. \ref{eq1}. In this way, for the classification tasks with large number of classes, one loop is enough to get the importance matrices for all classes.
\noindent As shown in the Fig. \ref{Fig. 4} (b), a direct and visual application of the importance matrix is that a binarized thresholded importance matrix can be applied to the CAM generation process as a purposeful filter, to filter out the features that are not frequently activated for the selected class or generate CAMs based only on principal features (e.g., Top 5\% features in the importance matrix). The revised CAM generation is given by:
\begin{equation}\label{eq5}
\begin{split}
FS-CAM=\sigma (scale_{space}&( im\cdot weights \\
&\cdot features))\\  
\end{split}
\end{equation}
\noindent where $im$ represents a mask based on the importance matrix, generated by the importance matrices being thresholded according to a specific purpose, while others remain the same. 
\noindent For example, when the importance matrices are thresholded to get the principal features, confounders or a single feature, a CAM based only on these selected features could be generated. Similarly to commonly scaled CAMs (indicated by prefix ‘S-‘), we indicate these “Feature Selected” CAMs computed from a mask by a prefix “FS-”For example, for S-Grad CAM, after the proposed masking, we call it “FS-S-Grad CAM”. The performance of the FS-CAMs compared to the original CAMs demonstrates the contribution of the selected features out of all features, and therefore the FS-CAMs for the principal features are used for the validation of the feature decomposition based on the importance matrix.

\section{Evaluation}
\noindent Since the proposed integrated feature analysis could contribute to improving the CAMs, we evaluated the proposed method by evaluating the corresponding CAMs compared to the baseline CAMs based on current class activation mapping algorithms. To evaluate the proposed methods, we introduce two metrics based on the CAMs and another metric based on the accuracy of models. The two metrics based on CAMs are based on the performance of CAMs in object localization and the consistency between the class activation levels of CAMs and models’ logits. The accuracy-based metric evaluates the feature decomposition according to the changes in model accuracy when different features are masked.
\subsection{Object localization}
\noindent The evaluation based on object localization and its variants, the most widely-used evaluation metrics, originate from the previous work \cite{GradCAMpp}, through measuring how much are the confidences affected by the regions highlighted in CAMs, using so-called “average increase” and “average drop (decrease)”. It starts by inputting the original image, and subsequently the original image is masked by the CAMs (thresholded and normalized) and fed into the model. This generates two output confidences (one from the original image and one from the masked image). Subsequently, the average increase and decrease (drop) in confidence due to the masking is calculated over the validation dataset. 
\noindent However, these evaluation metrics have a significant disadvantage: the changes can be caused by both  the model and the model interpretation algorithms, yet they are used to evaluate the model interpretation algorithms only, without excluding the impact of model performances and prior knowledge involvements. Evaluating interpretation algorithms through the accuracy of locating the target object also requires certain assumptions: (1) a perfect model that has perfect performance and is getting output classes “only” based on the target objects – without any inference from the environment or other objects that may correlate to the output classes. This assumption is so strict that no model today has been proven to meet this requirement. In addition, this evaluation also requires the model interpretation algorithms to have no prior knowledge that would lead to perfect object localization without models involved. In fact, a randomly generated model without training could receive “convincing” CAMs with an over-designed model interpretation algorithm according to \cite{SanityCheck}
\noindent Due to the above reasons, in this work, the average increase and decrease are only applied to prove that principal features according to the feature decomposition are enough for model to get the output classes and generate CAMs to locate objects, as it is indeed a good application in the field of weakly supervised object detection. 

\subsection{Consistency with the model’s logits}
\noindent In the Introduction section, we showed that class activation mapping algorithms assume that the well-weighted features can explain the model decision. Moreover, the class activation mapping algorithms also rely on the regions containing objects of the selected class being more activated than the other regions or images without objects belonging to the selected classes. Based on these facts, good CAMs should directly reflect the confidences/logits of the models using the class activation values at the dataset level. Therefore, we propose to evaluate the effectiveness of CAMs by calculating the correlation between the total class activation levels of the CAMs and the corresponding model confidences, which are independent of the model performance and any prior knowledge.
\noindent We use the correlation between logits and the sum of class activation levels in CAMs as the metric, using Pearson's correlation coefficient if they satisfy the bivariate normal distribution, and Spearman's rank correlation coefficient if they do not. Using logits instead of confidences may improve readability, as the activation functions (e.g. the SoftMax function) break the linear relationship between logits and confidences by rescaling according to the logits for other classes, and thus lead to a non-linear relationship between model logits and class activations.

\subsection{Accuracy changes after feature masking}
\noindent Using the difference in accuracy between the model with and without feature selection is a direct way to evaluate the contribution of the selected features. This masking method is different from the masking method used in the object localization evaluation and is applied to the feature instead of the input. Specifically, to validate the effectiveness and contribution of the selected features, we propose to block the forward paths of the unselected features during inference, obtain the logits based only on the selected features, and then take the accuracy changes as the evaluation metric. This could avoid the accuracy decreases caused by other factors, as masking the input can lead to the destruction of the input distribution and result in a significant decrease in accuracies.
\noindent We used the accuracy changes with feature masking to evaluate the effectiveness of feature decomposition.

\section{Materials}
\noindent Eight datasets and eight models from different research fields are used for experiments, together with eight different CAM algorithms applied for comparison and backbones of the integrated feature analysis.

\subsection{Datasets and models}
The datasets and corresponding models used for method validation include: 
\begin{itemize}
    \item ILSVRC2012 \cite{ILSVRC2012} for 1000-class natural objects, using standard ResNet18/34/50 and VGG11/13/16 as models, with top-1 accuracies ranging from 70\% to 83\%;
    \item Cats \& Dogs \cite{CatsDogs} for classifying cats and dogs, using standard ResNet18/34/50 and VGG11/13/16 as the models, achieving accuracies around 99.5\% (ResNet34) and 98.6\% (VGG16) on 10000 test images;
    \item MNIST \cite{MNIST} for classifying ten-class digits, using a model consisting of a simple two-layer multi-head self-attention block (from Transformers) with a multi-layer perceptron (fully connected layers), achieving an accuracy of over 99.9\% on test digits;
    \item MRI (T1-weighted contrast-enhanced with fat suppression) from the ESMIRA project \cite{ESMIRA} for classification of rheumatoid arthritis (RA). This (non-public) dataset, containing over 6000 3D MRIs from 2000 subjects, includes three classes: early arthritis clinic (EAC), clinically suspect arthralgia (CSA) and healthy controls (ATL). The task is to discriminate between these classes. The model used is a 2D plus 3D U-net encoder with a multilayer perceptron that achieves an AUC of 83\%;
    \item Cropped CT scans from the LIDC/IDRI malignancy detection database \cite{luna}, where the target objects are the malignant lesions. The model for this task achieved an AUC of 81\% and is a copy of VGG11;
    \item X-rays from the RSNA pneumonia detection task \cite{rsna}, the target objects are the regions with signs of pneumonia. The model for this task achieved an AUC of 84\% and is a copy of ResNet34;
    \item Ultrasound for breast cancer classification \cite{us}, and the target objects are the cancer-related regions. The model for this task achieved an AUC of 76\% and is a simple multilayer CNN described in the literature;
    \item Skin image from the SIIM-ISIC melanoma classification \cite{siim}; the target objects are the regions associated with malignant skin cancers. The model for this task achieved an AUC of 78\% and is a copy of VGG11 (resulting in low resolution due to down-sampling).
\end{itemize}

\begin{figure}[!t]
\centering
\includegraphics[width=\linewidth]{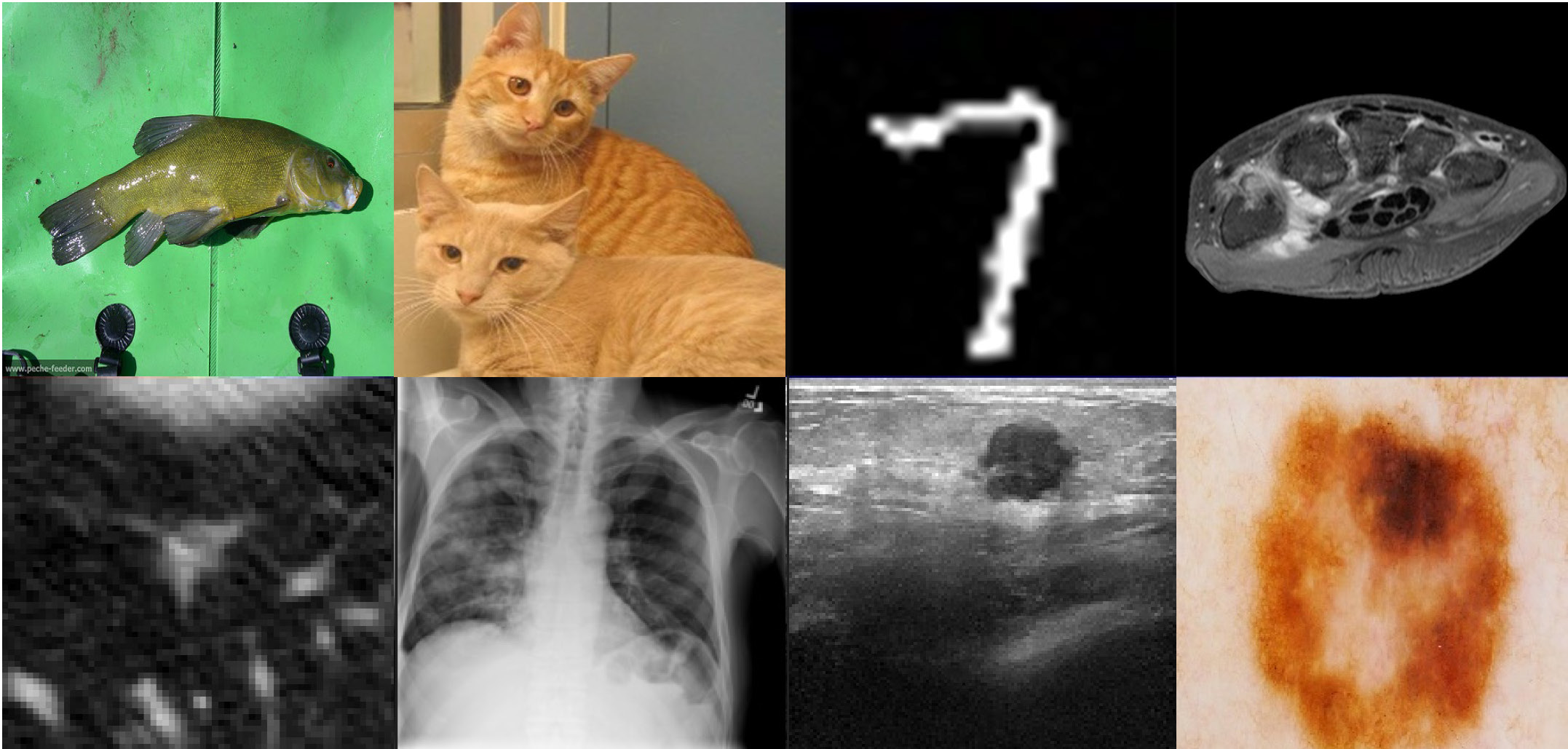}
\caption{Some examples from the image datasets used. Images from the top left to the bottom right are: a natural image from 1000 classes, a cat image from two-class Cats\&Dogs, a digit, an MRI, a CT, an X-ray, an ultrasound image and a skin photograph.}
\label{Fig. 5}
\end{figure}

\noindent The discussion about the types of models and input modalities can be found in the section of generalizability in the supplementary materials. 

\subsection{CAMs involved}
\noindent The CAM algorithms involved in the experiment are Grad CAM \cite{GradCAM}, Grad CAM++ \cite{GradCAMpp}, xGrad CAM \cite{XGradCAM}, Score CAM \cite{ScoreCAM}, Smooth Grad CAM++ \cite{SmoothGradCAMpp}, SS-CAM \cite{SSCAM}, IS-CAM \cite{ISCAM} and pixel-wise Grad CAM (pixel-wise gradients with features).
\noindent Since the branch of perturbation-based weights (Score CAM, SS-CAM, etc.) is unacceptably time-consuming for a broad quantitative evaluation, and because the purpose of this work is not to compare different CAM algorithms, the quantitative evaluation focuses on Grad CAM, Grad CAM++, xGrad CAM and pixel-wise Grad CAM. For the branch of perturbation-based weights, such as Score CAM, we present some visual results instead.

\section{Experiments and results}
\noindent This section contains three subsections, including (1) visual examples of the S-CAMs and FS-S-CAMs compared to some existing CAM algorithms; (2) the quantitative evaluation of the consistency improvement with the common intensity scaling based on the distribution analysis; and (3) the quantitative evaluation of the principal features selected by the importance matrix.

\subsection{Visual checks of the S-CAMs and FS-S-CAMs}
\begin{figure*}[!t]
\centering
\includegraphics[width=\textwidth]{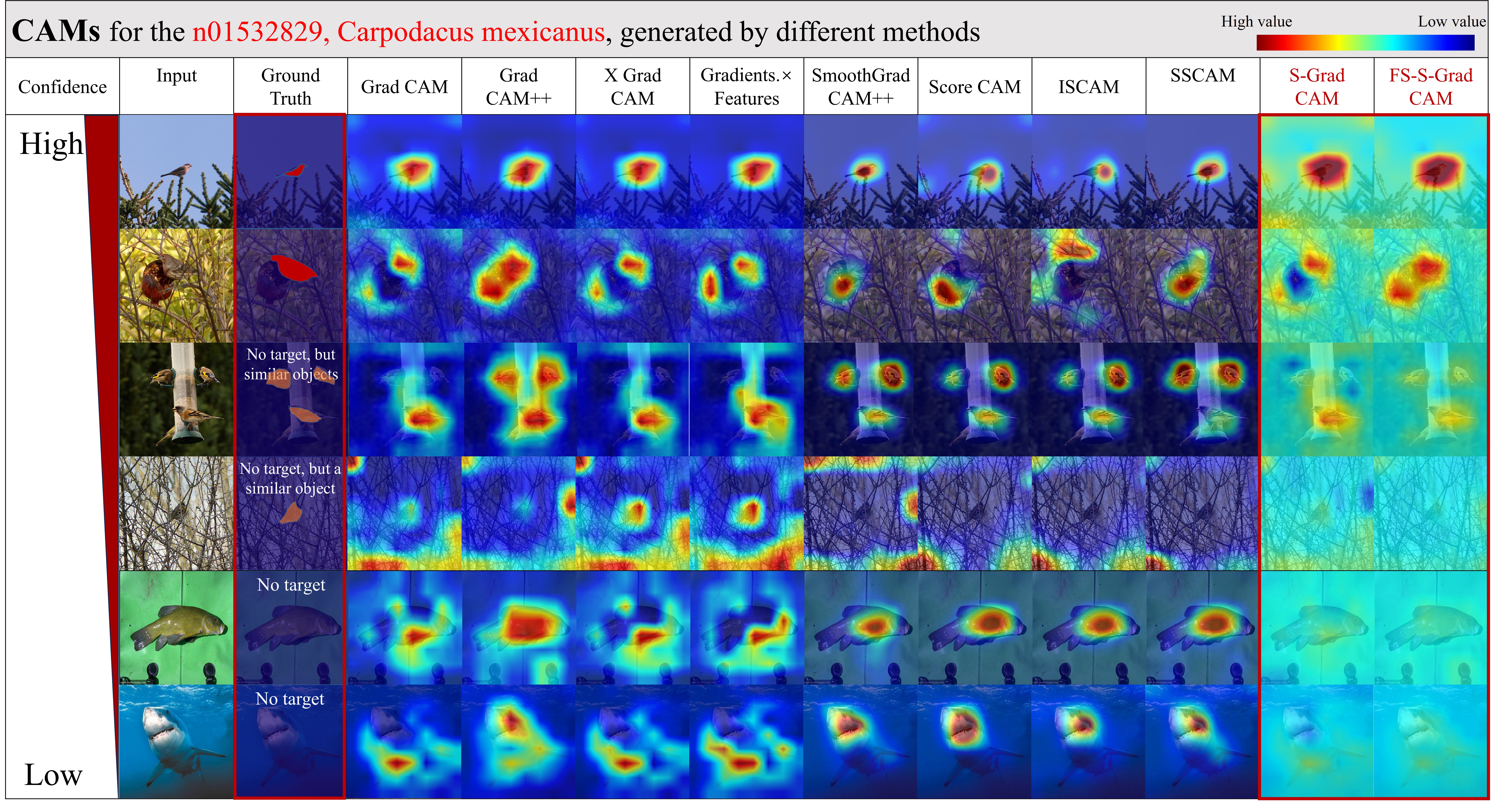}
\caption{The CAMs for a selected class of Carpodacus mexicanus, which are based on different CAM algorithms and the same ResNet34 model weights. With the decrease in confidences from the model, the CAMs are supposed to have decreasing class activation levels, yet the trends only appear in the S-Grad CAM and the FS-S-Grad CAM.}
\label{Fig. 6}
\end{figure*}

\begin{figure}[!t]
\centering
\includegraphics[width=\linewidth]{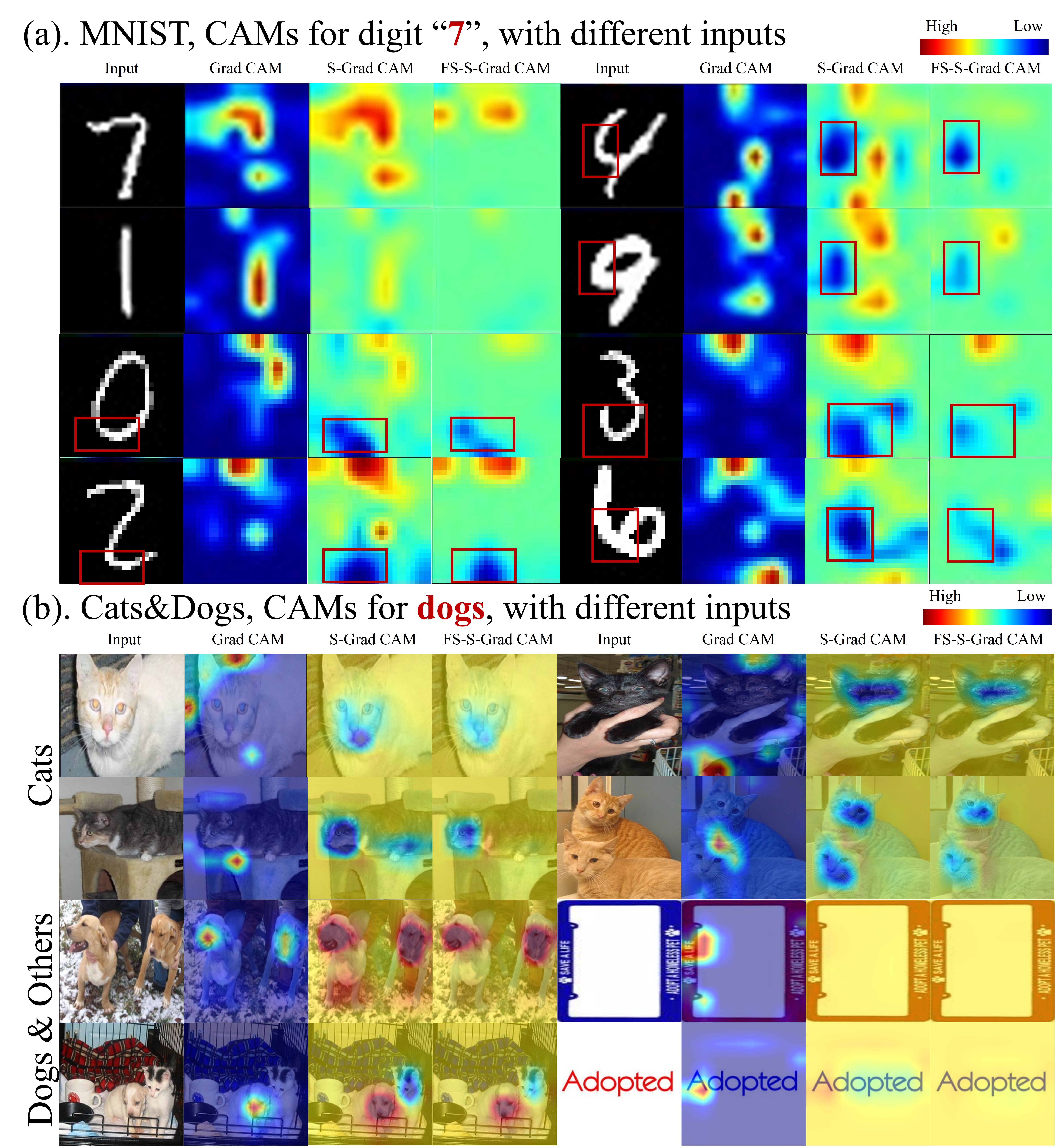}
\caption{Examples of CAMs for (a) MNIST (with a selected class of digit “7”), (b) Cats \& Dogs (with a selected class of “dogs”). In every row, the images are ordered as: input image, Grad CAMs, S-Grad CAMs, and the scaled CAMs with feature selection (FS-S-Grad CAM).}
\label{Fig. 7}
\end{figure}

\begin{figure}[!t]
\centering
\includegraphics[width=\linewidth]{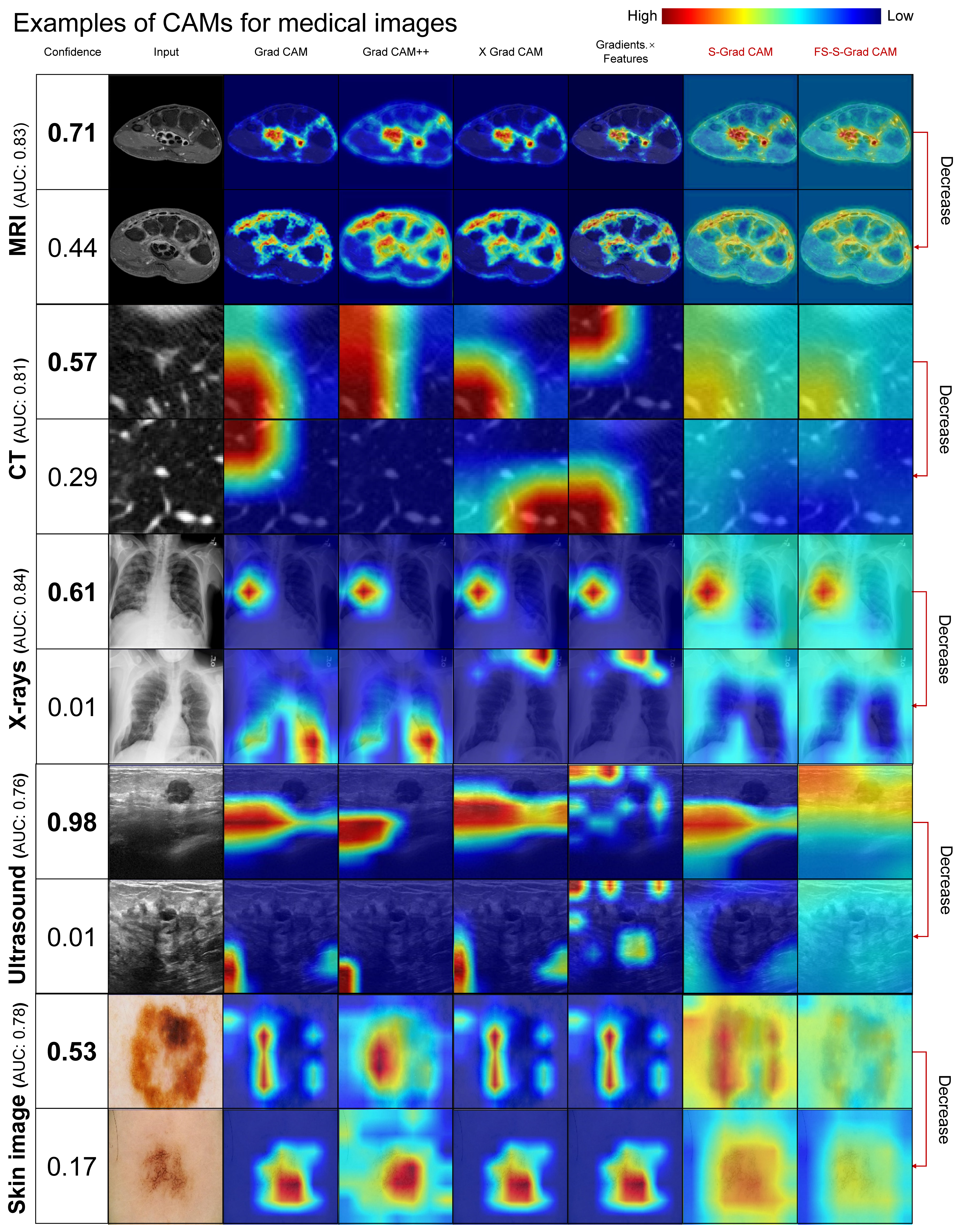}
\caption{Examples of CAMs for medical images with different confidences based on models. Please note that some saliency maps highlight regions without any visible objects or lesions and the models give output classes of diseases or lesions. This could happen for different reasons: (1) the models were not trained well; (2) the models found some confounders that were not visible; (3) the models found the images were “not normal enough” and overall tended to treat it as patients and put all those were not normal enough to patients; (4) the visualization method went wrong. In this figure, the S-Grad CAMs and FS-S-Grad CAMs show better consistency with the confidences, while CAMs based on other class activation mapping algorithms get more class activations from the input with lower confidences than the higher confidences.}
\label{Fig. 8}
\end{figure}

\noindent Fig. \ref{Fig. 6} to \ref{Fig. 8} briefly show some examples based on existing class activation mapping algorithms compared to the CAMs with the common intensity scaling based on the P90 and P10 from the distribution analysis. In the visual checks, we present the S-Grad CAM and FS-S-Grad CAM, which is based on the most basic CAM algorithm, to show the importance and effectiveness of the common intensity scaling and the results that it could exceed other CAM algorithms on matching the model confidences. 
\noindent Fig. \ref{Fig. 6} presents an example of the CAMs for a selected class of Carpodacus mexicanus (ImageNet2012) based on a standard ResNet34. From the perspective of matching the confidences, S-Grad CAM and FS-S-Grad CAM with common intensity scaling managed to show a decreasing trend on their CAMs as the confidences decreased, while other methods failed. 

\noindent Fig. \ref{Fig. 7} presents an example of the CAMs for a selected class of digit “7” \cite{MNIST} based on a model with a two-layer multi-head self-attention block followed by a multilayer perceptron, and an example of the CAMs for a selected class of “dogs” \cite{CatsDogs} based on a standard VGG16. In the CAMs for digit “7”, the red boxes indicate the regions that have negative contributions to the model’s decision to give an output class of “7” according to S-Grad CAM. This results fits our intuition since there should be nothing around the left and bottom left regions around the digit “7” and are supposed to have some intensities appearing on the top regions. In the CAMs for the selected class “dogs”, the S-Grad CAM did not only remove the random highlighted regions in the backgrounds when input was a cat or an undefined object. The S-Grad CAM gave class activation levels of less than the backgrounds for cats as they are the opposite class of dogs in this dataset, and gave overall “background” class activation levels to the undefined inputs.

\noindent Fig. \ref{Fig. 8} presents some examples of the CAMs with a selected class of lesion existence, inflammation areas or any signs of diseases (the opposite classes of normal/healthy) based on the five different modalities in medical imaging, using the models described in the Material section. These examples are shown to prove in medical images, that the same problem of inconsistency between class activations and model logits also exists in the original CAMs. In these applications, scalar outputs are more frequently required than in the previous datasets. In the medical domain scalar outputs (logits or confidences) should demonstrate the severity of diseases/symptoms or the chances of a developing disease. With a common intensity scale, medical images with higher model confidences on lesions, inflammation areas or diseases received more and higher class activations than those with lower model confidences.

\subsection{Quantitative evaluation: the common intensity scaling}
\begin{figure}[!t]
\centering
\includegraphics[width=\linewidth]{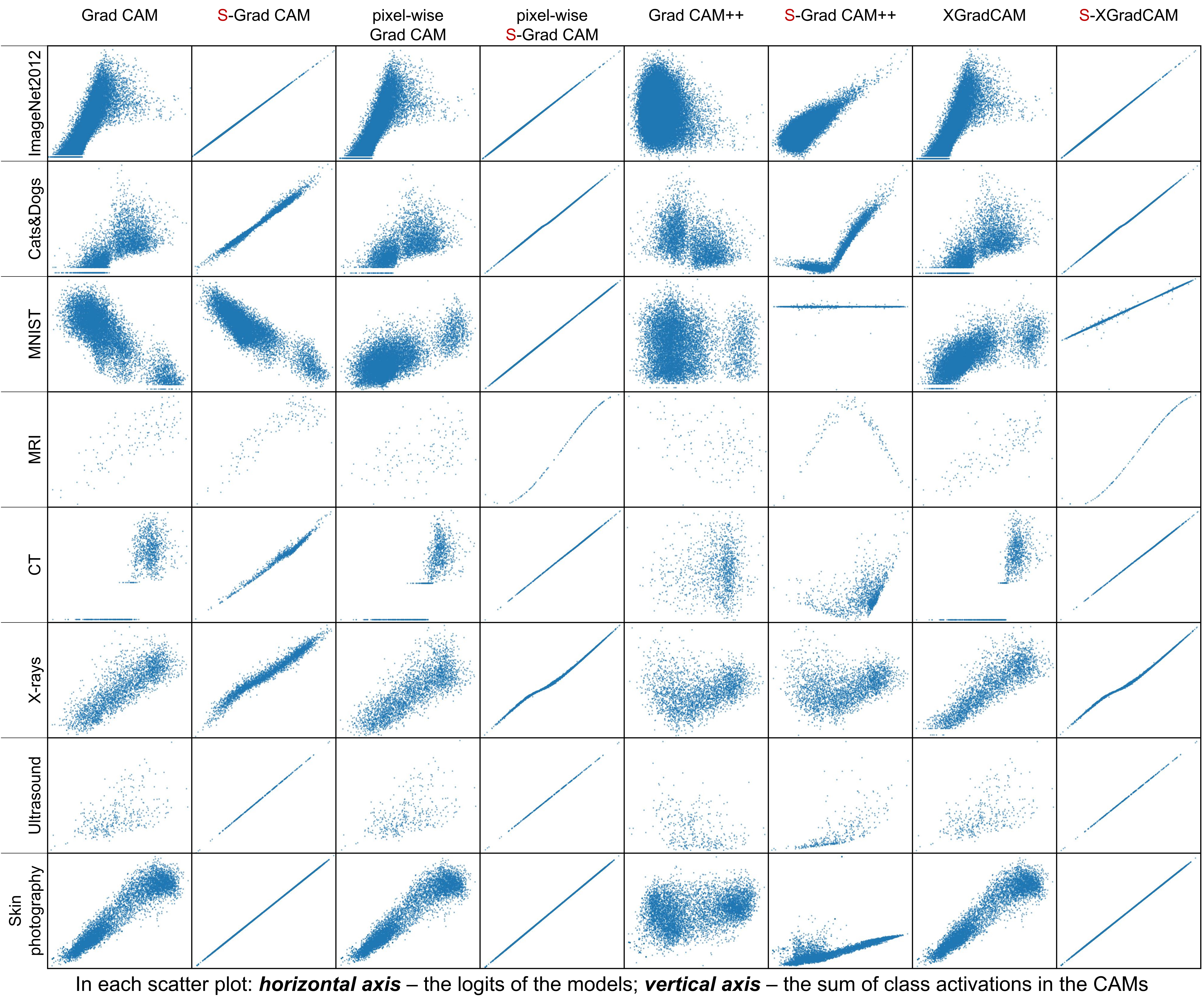}
\caption{The scatter plots for the sum of class activations in CAMs (vertical axis) and the logits of the models (horizontal axis) in each dataset, comparing the CAMs based on existing methods with individual intensity scales and the ones with common intensity scales. In most cases, the common intensity scaling significantly improved the correlation between CAMs and models.}
\label{Fig. 9}
\end{figure}

\noindent Fig. \ref{Fig. 9} presents some examples of the scatter plots that show the relation between the sum of class activations in CAMs and model logits. Table \ref{tab1} provides more details with the correlation coefficients from each dataset, based on the original CAM algorithms and with the proposed common intensity scale.

\noindent In Fig. \ref{Fig. 9}, most CAMs received high (nearly 100\%) correlation coefficients using the common intensity scale (especially for pixel-wise S-Grad CAM and S-XGrad CAM), however, the CAMs based on some class activation mapping algorithms may perform relatively worse than others. These errors originate from the weight definition, for example, for the scatter plots of MNIST-Grad CAM and MNIST-S-Grad CAM, the calculation of the weights of Grad CAM is based on the average of the gradients in the feature, which is not appropriate or MNIST dataset, as in this dataset, the digits appear more frequently in the center of the image.

\begin{table*}[]
\centering
\caption{The correlation coefficients of the original class activation mapping algorithms and the ones using a common intensity scale with the logits of the models. All results are statistically significant with p values less than 0.005.}
\begin{tabular}{|c|c|c|c|c|c|c|c|c|c|}
\hline
Corr.Coeff.      & \begin{tabular}[c]{@{}c@{}}Grad\\ CAM\end{tabular} & \begin{tabular}[c]{@{}c@{}}S-Grad \\ CAM\end{tabular} & \begin{tabular}[c]{@{}c@{}}pixel-wise \\ Grad CAM\end{tabular} & \begin{tabular}[c]{@{}c@{}}pixel-wise\\ S-Grad CAM\end{tabular} & \begin{tabular}[c]{@{}c@{}}Grad \\ CAM++\end{tabular} & \begin{tabular}[c]{@{}c@{}}S-Grad \\ CAM++\end{tabular} & XGradCAM & S-XGradCAM & Ave.Inc.(\%) \\ \hline
ImageNet2012     & 0.892                                              & 0.999                                                 & 0.891                                                          & 0.999                                                           & 0.048                                                 & 0.651                                                   & 0.892    & 0.999      & 23.1\%       \\ \hline
Cats\&Dogs       & 0.754                                              & 0.995                                                 & 0.686                                                          & 0.999                                                           & -0.422                                                & 0.820                                                   & 0.729    & 0.999      & 51.6\%       \\ \hline
MNIST            & -0.678                                             & -0.811                                                & 0.545                                                          & 0.999                                                           & 0.019                                                 & 0.029                                                   & 0.708    & 0.998      & 15.5\%       \\ \hline
MRI              & 0.704                                              & 0.763                                                 & 0.457                                                          & 0.999                                                           & -0.331                                                & -0.271                                                  & 0.752    & 0.999      & 22.7\%       \\ \hline
CT               & 0.553                                              & 0.980                                                 & 0.758                                                          & 0.999                                                           & 0.071                                                 & 0.485                                                   & 0.765    & 0.999      & 32.9\%       \\ \hline
X-rays           & 0.849                                              & 0.983                                                 & 0.827                                                          & 0.999                                                           & 0.292                                                 & 0.292                                                   & 0.874    & 0.999      & 10.7\%       \\ \hline
Ultrasound       & 0.532                                              & 1.000                                                 & 0.532                                                          & 1.000                                                           & -0.340                                                & 0.825                                                   & 0.532    & 1.000      & 64.2\%       \\ \hline
Skin photography & 0.946                                              & 0.999                                                 & 0.945                                                          & 0.999                                                           & 0.266                                                 & 0.790                                                   & 0.945    & 0.999      & 17.1\%       \\ \hline
\end{tabular}
\label{tab1}
\end{table*}

\subsection{Quantitative evaluation: the selected principal features}
\noindent Table. \ref{tab2} presents the accuracies from the models using all features and using only the selected principal features (top 5\% to 25\% based on the feature decomposition). Furthermore, Fig. 10and Table. 3 present the two different ways of evaluating the effectiveness of the selected principal features: the CAMs based only on these selected features could reach (1) the consistency with the logits of the models and (2) the performance in object localization with the models as good as the CAMs using all features. 
\noindent In all these approaches, the effectiveness of the feature decomposition is validated by evaluating the principal features selected based on the importance matrices. Based on the average class activations of each feature in importance matrices, we took the top 5\% (top 25\% ImageNet and top 20\% for MNIST) features in the importance matrices as the principal features and used them to conduct the experiments.

\begin{table}
\centering
\caption{The accuracies of the models inferring on eight datasets, based on top 5\% features and all features available at the selected layers of the models.}
\resizebox{\linewidth}{!}{
\begin{tabular}{|c|c|c|c|}
\hline
Dataset                                                                & \begin{tabular}[c]{@{}c@{}}Acc.\\ (All features)\end{tabular} & \begin{tabular}[c]{@{}c@{}}Acc.\\ (Principal features)\end{tabular} & \begin{tabular}[c]{@{}c@{}}Acc.\\ (Non-principal features)\end{tabular} \\ \hline
\begin{tabular}[c]{@{}c@{}}ImageNet2012\\ (Natural image)\end{tabular} & 0.706                                                         & 0.689                                                               & 0.034                                                                   \\ \hline
\begin{tabular}[c]{@{}c@{}}Cats\&Dogs\\ (Natural image)\end{tabular}   & 0.985                                                         & 0.962                                                               & 0.507                                                                   \\ \hline
\begin{tabular}[c]{@{}c@{}}MNIST\\ (Digit)\end{tabular}                & 0.994                                                         & 0.933                                                               & 0.075                                                                   \\ \hline
\begin{tabular}[c]{@{}c@{}}ESMIRA\\ (MRI)\end{tabular}                 & 0.831                                                         & 0.772                                                               & 0.532                                                                   \\ \hline
\begin{tabular}[c]{@{}c@{}}CT\\ (cropped LIDC/IDRI)\end{tabular}       & 0.813                                                         & 0.772                                                               & 0.524                                                                   \\ \hline
\begin{tabular}[c]{@{}c@{}}X-rays\\ (RSNA pneumonia)\end{tabular}      & 0.839                                                         & 0.797                                                               & 0.506                                                                   \\ \hline
\begin{tabular}[c]{@{}c@{}}Ultrasound\\ (usbc)\end{tabular}            & 0.764                                                         & 0.642                                                               & 0.511                                                                   \\ \hline
\begin{tabular}[c]{@{}c@{}}Skin photography\\ (SIIM-ISIC)\end{tabular} & 0.78                                                          & 0.735                                                               & 0.525                                                                   \\ \hline
\end{tabular}}
\label{tab2}
\end{table}

\noindent Table. \ref{tab2} presents the accuracies using only selected features to infer the output classes compared to using all features available at the selected layers of the models. Except for the model on ultrasound, the selected principal features succeed in containing the most important information needed for the model, achieving very close accuracies as the original inference using all features available at the selected layers of the models. For ultrasound, the top 5\% features are not sufficient, while the top 25\% and 20\% features are necessary for ImageNet and MNIST to cover most of the information required for model to get the correct output classes.

\begin{figure}[!t]
\centering
\includegraphics[width=\linewidth]{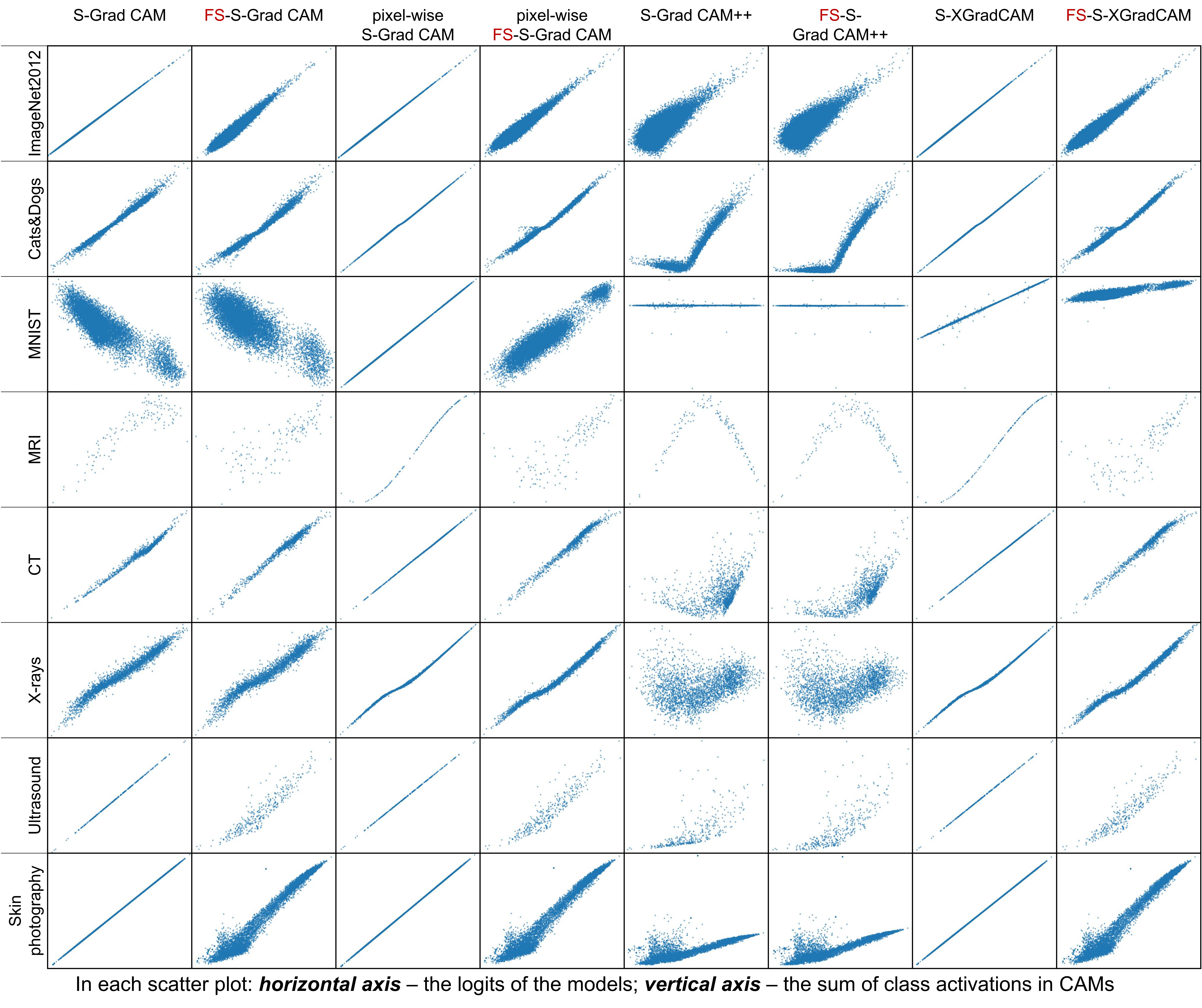}
\caption{The scatter plots for the sum of class activations in CAMs and the logits of the models in each dataset, comparing the CAMs based on all features available at the selected layers of the models and on principal features (top 5\% for six datasets, 20\% for MNIST and 25\% for ImageNet2012). Although the drop of most features leads to losses in consistency, the class activations of the CAMs based only on principal features maintained a high level of consistency with the logits of the models.}
\label{Fig. 10}
\end{figure}

Fig. \ref{Fig. 10} gives the performance on the consistency evaluation, which demonstrates that 5\% of the features selected by importance matrices indeed contain most information needed for the models to infer the output classes, and could achieve similar consistency with the models as compared to using all features at the selected layers of the models. Similarly, from the Table. \ref{tab3}, the accuracy dropped only gradually when selecting only the top 5\% (20\% for MNIST and 25\% for ImageNet) features: less than 10\% in most datasets. Furthermore, the consistency between the CAMs and the model’s logits could even increase in some situations, by filtering out the less-activated features.

\begin{table*}
\centering
\caption{The correlation coefficients of the CAMs and the logits of the models, comparing the S-CAMs and FS-S-CAMs.}
\resizebox{\textwidth}{!}{
\begin{tabular}{|c|c|c|c|c|c|c|c|c|c|}
\hline
Corr.Coeff.      & S-Grad CAM & \begin{tabular}[c]{@{}c@{}}FS-S-Grad \\ CAM\end{tabular} & \begin{tabular}[c]{@{}c@{}}pixel-wise\\ S-Grad CAM\end{tabular} & \begin{tabular}[c]{@{}c@{}}pixel-wise \\ FS-S-Grad CAM\end{tabular} & \begin{tabular}[c]{@{}c@{}}S-Grad \\ CAM++\end{tabular} & \begin{tabular}[c]{@{}c@{}}FS-S-Grad \\ CAM++\end{tabular} & \begin{tabular}[c]{@{}c@{}}S-XGrad\\ CAM\end{tabular} & \begin{tabular}[c]{@{}c@{}}FS-S-\\ XGradCAM\end{tabular} & Ave.Dec.(\%) \\ \hline
ImageNet2012     & 0.999      & 0.915                                                    & 0.999                                                           & 0.914                                                               & 0.651                                                   & 0.652                                                      & 0.999                                                 & 0.915                                                    & 6.35\%       \\ \hline
Cats\&Dogs       & 0.995      & 0.992                                                    & 0.999                                                           & 0.993                                                               & 0.820                                                   & 0.871                                                      & 0.999                                                 & 0.992                                                    & -0.88\%      \\ \hline
MNIST            & -0.811     & -0.703                                                   & 0.999                                                           & 0.728                                                               & 0.029                                                   & 0.049                                                      & 0.998                                                 & 0.728                                                    & 11.33\%      \\ \hline
MRI              & 0.763      & 0.797                                                    & 0.999                                                           & 0.849                                                               & -0.271                                                  & -0.143                                                     & 0.999                                                 & 0.856                                                    & 3.28\%       \\ \hline
CT               & 0.980      & 0.972                                                    & 0.999                                                           & 0.981                                                               & 0.485                                                   & 0.752                                                      & 0.999                                                 & 0.981                                                    & -5.57\%      \\ \hline
X-rays           & 0.983      & 0.978                                                    & 0.999                                                           & 0.995                                                               & 0.292                                                   & 0.355                                                      & 0.999                                                 & 0.996                                                    & -1.27\%      \\ \hline
Ultrasound       & 1.000      & 0.927                                                    & 1.000                                                           & 0.927                                                               & 0.825                                                   & 0.829                                                      & 1.000                                                 & 0.927                                                    & 5.37\%       \\ \hline
Skin photography & 0.999      & 0.931                                                    & 0.999                                                           & 0.931                                                               & 0.790                                                   & 0.815                                                      & 0.999                                                 & 0.931                                                    & 4.47\%       \\ \hline
\end{tabular}}
\label{tab3}
\end{table*}

\begin{table}
\centering
\caption{The changes of the average increases and decreases (drops) based on difference CAM algorithms, comparing the original CAMs and FS-CAMs. For average increase, the higher the better; for average decrease, the lower the better. Percentage changes in metrics were applied to increase readability.}
\resizebox{\linewidth}{!}{
\begin{tabular}{|c|c|c|c|c|c|}
\hline
Change of metrics                                                  & \begin{tabular}[c]{@{}c@{}}Grad\\ CAM\end{tabular} & \begin{tabular}[c]{@{}c@{}}pixel-wise \\ Grad CAM\end{tabular} & \begin{tabular}[c]{@{}c@{}}Grad\\ CAM++\end{tabular} & \begin{tabular}[c]{@{}c@{}}XGrad\\ CAM\end{tabular} & Average \\ \hline
\begin{tabular}[c]{@{}c@{}}Increase \\ (ImageNet2012)\end{tabular} & -6.01\%                                            & -11.16\%                                                       & 1.85\%                                               & -4.20\%                                             & -4.88\% \\ \hline
\begin{tabular}[c]{@{}c@{}}Decrease \\ (ImageNet2012)\end{tabular} & 2.67\%                                             & 6.44\%                                                         & -1.50\%                                              & 2.04\%                                              & 2.41\%  \\ \hline
\begin{tabular}[c]{@{}c@{}}Increase \\ (Cats\&Dogs)\end{tabular}   & -6.22\%                                            & -10.68\%                                                       & -7.41\%                                              & -11.02\%                                            & -8.83\% \\ \hline
\begin{tabular}[c]{@{}c@{}}Decrease \\ (Cats\&Dogs)\end{tabular}   & 9.09\%                                             & 23.81\%                                                        & 8.33\%                                               & 19.05\%                                             & 15.07\% \\ \hline
\begin{tabular}[c]{@{}c@{}}Increase \\ (MNIST)\end{tabular}        & 100.00\%                                           & -33.33\%                                                       & 0.00\%                                               & 0.00\%                                              & 16.67\% \\ \hline
\begin{tabular}[c]{@{}c@{}}Decrease \\ (MNIST)\end{tabular}        & -24.02\%                                           & 0.55\%                                                         & 12.62\%                                              & 11.49\%                                             & 0.16\%  \\ \hline
\begin{tabular}[c]{@{}c@{}}Increase \\ (MRI)\end{tabular}          & -1.85\%                                            & -4.07\%                                                        & -1.57\%                                              & -1.47\%                                             & -2.24\% \\ \hline
\begin{tabular}[c]{@{}c@{}}Decrease \\ (MRI)\end{tabular}          & -1.42\%                                            & 1.79\%                                                         & 4.66\%                                               & -2.16\%                                             & 0.72\%  \\ \hline
\begin{tabular}[c]{@{}c@{}}Increase \\ (CT)\end{tabular}           & -5.73\%                                            & 0.37\%                                                         & -5.92\%                                              & -3.52\%                                             & -3.70\% \\ \hline
\begin{tabular}[c]{@{}c@{}}Decrease \\ (CT)\end{tabular}           & 10.20\%                                            & 1.83\%                                                         & 4.83\%                                               & 10.83\%                                             & 6.92\%  \\ \hline
\begin{tabular}[c]{@{}c@{}}Increase \\ (X-rays)\end{tabular}       & -0.79\%                                            & -7.14\%                                                        & 10.43\%                                              & -7.32\%                                             & -1.20\% \\ \hline
\begin{tabular}[c]{@{}c@{}}Decrease \\ (X-rays)\end{tabular}       & 1.20\%                                             & 8.88\%                                                         & -1.20\%                                              & 8.45\%                                              & 4.33\%  \\ \hline
\begin{tabular}[c]{@{}c@{}}Increase \\ (Ultrasound)\end{tabular}   & 117.95\%                                           & 0.00\%                                                         & -61.54\%                                             & -12.90\%                                            & 10.88\% \\ \hline
\begin{tabular}[c]{@{}c@{}}Decrease \\ (Ultrasound)\end{tabular}   & 1.67\%                                             & -5.21\%                                                        & 0.00\%                                               & 4.10\%                                              & 0.14\%  \\ \hline
\begin{tabular}[c]{@{}c@{}}Increase \\ (Skin)\end{tabular}         & -2.87\%                                            & -2.87\%                                                        & -0.63\%                                              & -2.87\%                                             & -2.31\% \\ \hline
\begin{tabular}[c]{@{}c@{}}Decrease \\ (Skin)\end{tabular}         & 0.79\%                                             & 0.79\%                                                         & 0.80\%                                               & 0.79\%                                              & 0.79\%  \\ \hline
\end{tabular}}
\label{tab4}
\end{table}

\noindent Table. \ref{tab4} presents the results on object localization, using percentage changes in the metrics to increase readability. On average, the information losses for object localization are less than 10\%, and sometimes the models performs better with 5\% principal features. Full results of the increases and decreases can be found in the supplementary materials. The result proves that with only the 5\% features, we could generate CAMs that are enough in these applications to detect the target objects as precisely as using all features. The CAMs based only on principal features achieved as good performance as the CAMs using all features, proving the independency of the features and the reliability of feature decomposition in the integrated feature analysis.
\noindent We also introduced another way of defining and calculating the average increase and decrease, providing more information on the mechanism of the masking-based object localization, the details and results are presented in the supplementary materials, as well as the reasons of using the original CAMs instead of the CAMs with a common scale.

\section{Discussion}
\noindent In this paper, we provide a model interpretation method called the integrated feature analysis, based on the same premise as the CAM algorithms. It is not a new class activation mapping algorithm or a new definition of weights for CAMs, but a different perspective of using existing weighted features. We proved the reliability of the method, based on the consistency between CAMs using common intensity scales and models, and the effectiveness of the principal features. The intensity scaling and feature selection could be applied to most CAM methods without increasing computation time during inference, serving as a useful functional supplement. On the other hand, the value of the integrated feature analysis is more than improving the CAM methods. In the following subsections, we discuss about the uses of the importance matrix, generalizability of the proposed method and the definition of importance. Some other topics (e.g. regression tasks, time costs, and evaluation metrics) can be found in supplementary materials.

\subsection{Uses of the importance matrix}

\begin{figure*}[!t]
\centering
\includegraphics[width=\textwidth]{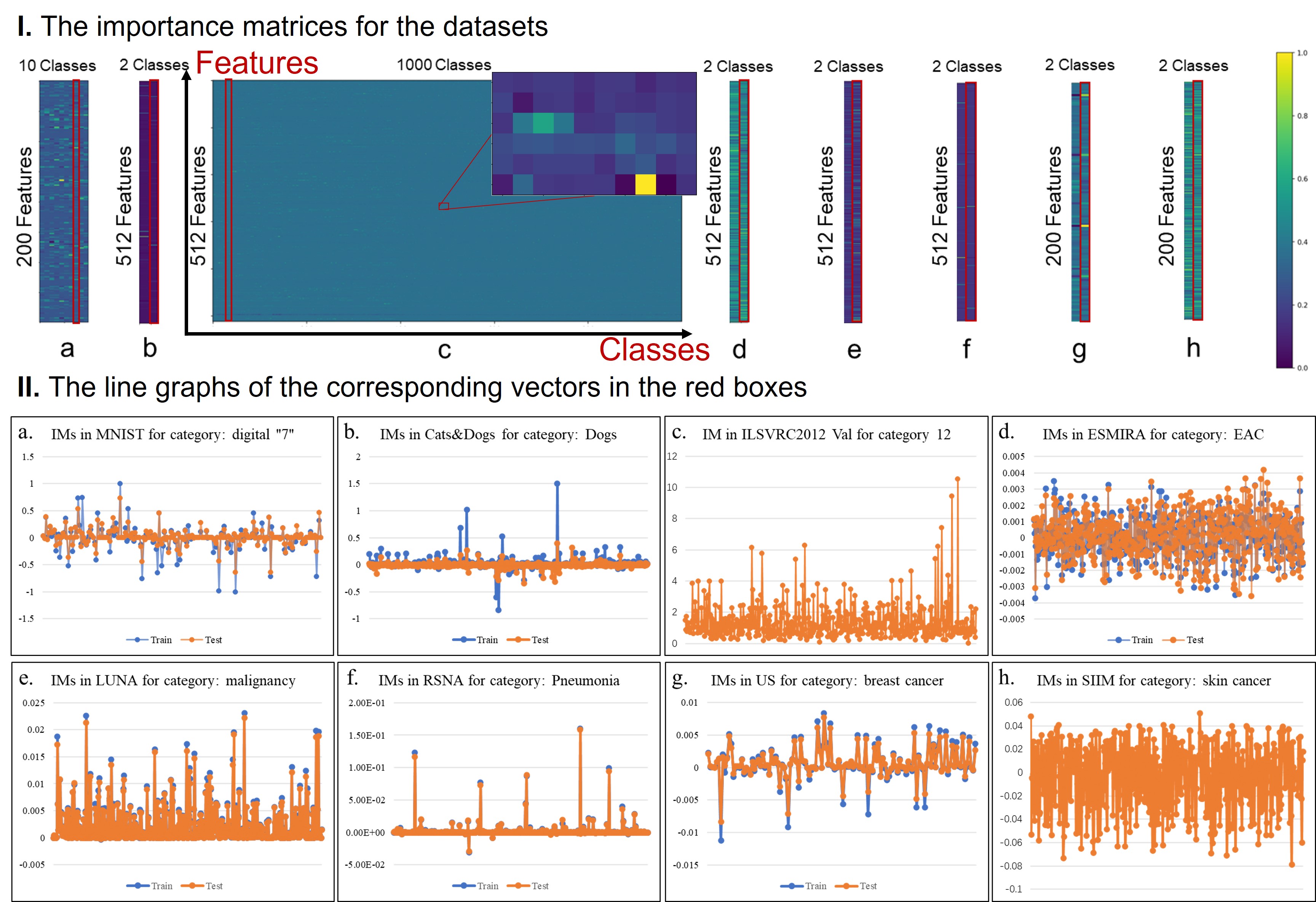}
\caption{(I) The importance matrices (resized for better visualization) for models trained on: (a) MNIST, (b) Cats \& Dogs, (c) ILSVRC2012 validation set only, (d) ESMIRA dataset when targeting the EAC, (e) Lung Nodule Analysis 2016 (LUNA) for malignancy (based on lesions), (f) RSNA Pneumonia Detection Challenge for pneumonia existence, (g) the breast ultrasound images dataset, and (h) SIIM-ISIC melanoma classification when targeting the melanomas. (II) The line graphs of the columns (red boxes in (I)) in the importance matrices (IMs), targeting one class and subtracting the mean level. The values represent the overall importance of certain features to the model decision, the line graphs of importance matrices based on the training set are shown in blue, and the ones based on the test set are shown in orange.}
\label{Fig. 11}
\end{figure*}

\noindent As the effectiveness of the feature decomposition has been proved by the experiments with principal features, the importance matrix indeed contains feature-wise information over the whole dataset and could be applied to the training set or the test set for different purposes. The importance matrix could provide useful information about both the model and dataset, such as:
\begin{itemize}
    \item Overfitting. The difference in the importance matrices between the training and test sets indicates potential overfitting when the class activations of each feature in the training set differ significantly from those in the test set. For example, in Fig. \ref{Fig. 11} (a), (b), (d), (e), (f) and (g) we show the importance matrices from both the training and test sets. From the difference between the training and test sets, (a) and (b) indicate more severe overfitting, as they have features that are much more activated in the training sets than in the test sets, while the training and test sets share a very similar pattern in other line graphs.
    \item Principal features. If we select a single column in the importance matrix (thereby selecting a class), the values in each row represent the average class activation level of each feature for that selected class. Those features that are more activated (higher class activations) also have a greater influence on this selected model output class in general. Therefore, they can serve as the main features of this study. For example, in the line graphs in Fig. \ref{Fig. 11}, except for (d) and (h), some features are more frequently activated with significant values, resulting in a large ratio of total activation values to average importance over the entire dataset. The values of these features indicate the potential of some features to emerge from many others.
    \item Potential confounders. The confounders in deep learning models usually refer to some objects or noise in the background of images or time series that could provide a "shortcut" for a model to be easily trained and make inferences based only on these non-targets. In the importance matrices, confounders could cause extremely high values for some specific features and relatively low values for all other features because they provide the "shortcut". These potential confounders can therefore be checked by generating CAMs using only the relevant features. As the models are copies of successful models on these public datasets, we did not find any significant signs of confounding in these datasets. Fig. \ref{Fig. 11} (f) may benefit from some potential confounders compared to other datasets, some visual checks on the CAMs are valuable for further investigation.
    \item Special cases and outliers. Special cases or outliers would show very different class activations of the features compared to the importance matrices of their corresponding classes. For example, in the blue boxes of Fig. \ref{Fig. 3}, we present a simple example from (d) of the ESMIRA project, where the sample belongs to the class "EAC", but has a different picture of activated features and was classified as the opposite with confidences around 0.5.
    \item Model redundancies. The ratio of highly activated features out of all features in the importance matrices represents the ratio of influential features in the model, as the values are a combination of activation frequency and "amplitude". Therefore, the rarely activated features in all classes/outputs contribute very little to the model decision in most cases and could be removed by model pruning. For example, in Fig. \ref{Fig. 11} (b) and (f), compared to other importance matrices, they have more features that are rarely activated across the datasets, while some features are significantly activated. This phenomenon indicates that the models have some "lazy" features that contribute very little and can be pruned.
\end{itemize}

\subsection{Generalizability of the feature analysis}
\noindent Although the proposed methods are mostly applied to convolutional neural networks, the importance matrix and the common intensity scaling are not limited to these type of networks. They are also feasible for other deep learning models, because the original CAM algorithms are also feasible for different types of models \cite{jacob-pytorch-cam-evaluation}. We had applied the proposed method to the tiny transformer on MNIST to prove its generalizability. However, the integrated feature analysis faced a transition problem that requires further studies on the definition of features in transformers and recurrent neural networks and the spatial reorganization to generate CAMs in these models.

\subsection{Definition of the importance}
\noindent For the importance matrices in this paper, we took the average of the sum of class activations over the whole dataset as the “importance”. However, the definition of the importance of each feature could be further investigated, as the standard deviation of the average of the weighted features could also be informative. Rarely activated features with massive class activation values in very few cases might obtain high values in the importance matrices based on the current definition. The definition of feature importance remains an open question, just as the definition of the weights for CAMs.

\section{Conclusion}
\noindent In summary, we proposed the integrated feature analysis for model interpretation by collecting the distribution information and decomposing the weighted features. It helps to improve the consistency between CAMs and models by constructing a common intensity scale and to obtain more information (e.g., main features) about models and datasets by constructing the importance matrix. As the experiments validate its effectiveness, and consider the potential uses of the importance matrix in analyzing models and datasets, integrated feature analysis appears to be a current "local-optimal" step towards better model interpretation and more informative class activation maps.

\bibliographystyle{unsrt}
\bibliography{ifaref}



\end{document}